# A FRAMEWORK FOR FAST SCALABLE BNN INFERENCE USING GOOGLENET AND TRANSFER LEARNING

KARTHIK E
SCHOOL OF ENGINEERING AND TECHNOLOGY,
CHRIST (DEEMED TO BE UNIVERSITY)
Karthik.e@btech.christuniversity.in
Karthikg098@gmail.com

# ABSTRACT

Efficient and accurate object detection in video and image analysis is one of the major beneficiaries of the advancement in computer vision systems with the help of deep learning. With the aid of deep learning, more powerful tools evolved, which are capable to learn high-level and deeper features and thus can overcome the existing problems in traditional architectures of object detection algorithms. The huge amount of visual information and demand of high throughput requirements are the major the bottle neck in the existing object detection architectures. The work in this project aims to achieve high accuracy in object detection with good real time performance.

In the area of computer vision, lot of research are going in the area of detection and processing of visual information, by improving the existing algorithms. Binarized neural network have shown high performance in various vision tasks such as image classification, object detection and semantic segmentation. In this thesis, object detection is done using binarized neural network with tensor flow to improve the accuracy and to reduce the execution time. The Modified National Institute of Standards and Technology database (MNIST), Canadian Institute For Advanced Research (CIFAR) and Street View House Numbers (SVHN) datasets are used which is implemented using pretrained convolutional neural network (CNN) that is 22 layers deep. Supervised learning is used in the work, which classifies the particular dataset with the proper structure of the model. In still images, to improve the accuracy, Googlenet is used. The final layer of the Googlenet are replaced with the transfer learning to improve the accuracy of the Googlenet. At the same time, the accuracy in moving images can be maintained by transfer learning techniques. Hardware is the main backbone for any model to obtain faster results with large number of datasets. Here, Nvidia Jetson Nano is used which is a graphics processing unit (GPU), that can handle large number of computations in the process of object detection. Results show that the accuracy of object detected by transfer learning method is more when compared to the existing methods.

# KEYWORDS

| Item | Description |
| --- | --- |
| **Framework** | Framework is an interface which allows to build deep learning models quickly into underlying algorithm. |
| **Googlenet** | Googlenet is a pre trained version of convolutional neural network that is 22 layers deep |
| **Transfer Learning** | Transfer Learning is used to store knowledge that is gained while solving one problem and applying it to a different but related problem. |
| **TensorFlow** | TensorFlow is an open source library which uses numerical computations to determine data flow graphs for solving mathematical operations. |
| **BNN** | Binarized **N**eural **N**etwork |
| **CNN** | Convolutional **N**eural **N**etwork |
| **GPU** | Graphics Processing Unit |
| **FPS** | Frames Per Unit |
| **BN** | Batch Normalization |

# INTRODUCTION

For humans, visual recognition is one of the five inevitable basic senses. In order to pick up a pen we need to find which part of object is recognized and later on grasp it. To recognize human, we determine where the face is a part of it [1]. For coordinating all these activities, the neurons in the brain sends the signal. Similarly, the deep learning techniques are used which mimics the neurons to identify an object. Several algorithms are developed for deep learning techniques but it's hardware implementation is the biggest challenge. First, the neural network model is trained and then with the help of computer vision the object is identified. For example, imagine a robot wants to pick a glass from the table. To perform this action, the robot must first find the glass and then the image classification is done to accurately finds the object [2].

Recently, the visual data has grown drastically and the size of the data to be processed by the algorithm is also huge. Now, nearly 1 billion photos are uploaded each day on Facebook [3], which is a typical example of the case. Besides this, the difficulty in classifying a set of grains into fine and coarse grains in unfavorable lighting condition also need to be addressed. With the advent of deep learning techniques, all the above challenges are effectively overcomed. Deep learning is sub-field of machine learning that studies statistical model called deep neural networks. Deep learning has made a big impact on computer vision which enhanced image classification and object detection. All these advantages come at the price of increase in computational resources becomes of its complex frameworks like Tensor flow and Torch7.

Graphics processing unit (GPU) is employed which increases the shared models and open source codes. Larger and deeper architecture are trained to handle with bigger datasets. Weakly supervised learning methods are often applied to beat the restrictions.

As different objects are on different scale of the pictures, therefore the window is employed to seek out all positions of the pictures. For an outsized dataset the window is difficult to use except for fixed images window method is suitable [5]. To acknowledge different objects feature extraction is applied. Manually it can distinguish all types of objects. Googlenet and TensorFlow is employed to spot the pictures from the dataset. A classifier distinguishes the target object to be identified. Among classifiers, the back- ground subtraction model [9] is employed which is more efficient. The goal of the work is to enhance BNN based real time object detection using Googlenet and transfer learning.

## MATERIAL AND METHODS

Binary neural network is difficult to train so the deep learning framework are used for training the networks. Googlenet dataset is of 22 layers were deep networks integrate high level features and classifiers. Network depth increases and accuracy gets saturated which degrades rapidly. It is not caused by over fitting model which leads higher training error. There exists a solution to deeper model the added layers are identifying mapping and other layers are of shallower model. Comprehensive experiments on Googlenet show that extremely deep nets are easy to optimize but exhibit higher training error when depth increases where deep nets can enjoy accuracy gain from increased depth [10].

The MNIST dataset has 95.8% accuracy where as CIFAR-10 and SVHN have 80.1% and 94.9% accuracy. CNN will require millions of floating-point operations are in tera operations per second on FPGA. CNN is focused on supervised learning and it receives input from small receptive field in the previous layer. Each pixel in the output image is the sum of all synapse weight and corresponding images. Pooling layers are simple down sample of 2D images. Max pooling replaces into small subtiles and replaces each subtile with largest element. Three types of binarizations are considered as binary input activations, binary synapse weight and binary output activations. The predetermined portion of synapse have zero weight and all other synapse have weight one 98.7% accuracy is determined by MNIST dataset. Systolic array is used which is a single processing engine style architecture using theoretical roofline model optimized for execution of each layer.

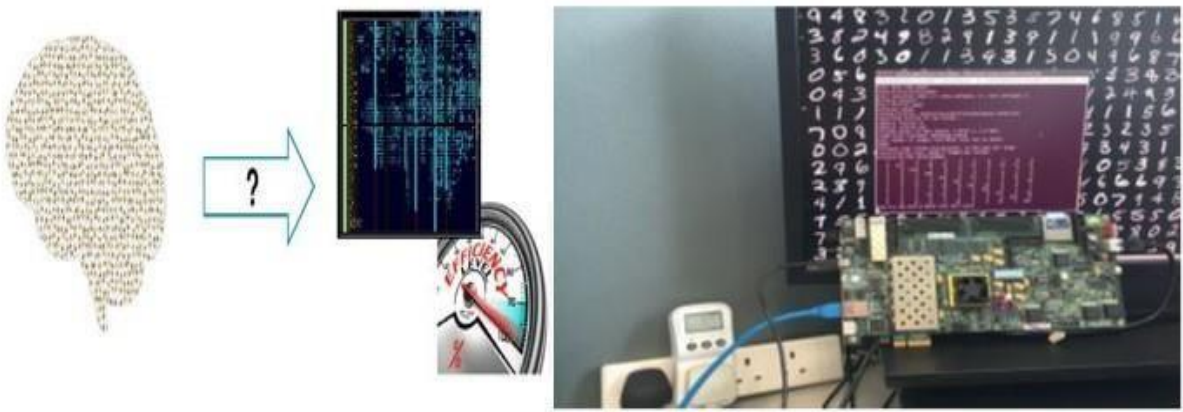

FIGURE 1: Implementation on Zynq board

Finn, a framework for building fast and flexible FPGA accelerator. CNNs implementations use floating point parameters. BNN performance using Roofline model with Alexnet is implemented on Zynq board. Trained using 32-bit Floating point numbers. The major advantages are comparing different platforms is to simply compare their accuracy, FPS and power consumption when working on the same benchmark datasets (MNIST, CIFAR-10 and SVHN) [7]. The disadvantage is all prototypes have been implemented on the Xilinx Zynq-7100.

**Comparison of Existing Problem Statements:**

| Research Papers- | Paper 1 [8] | Paper 2 [9] | Paper 3 [10] |
| --- | --- | --- | --- |
| Implementation | Implemented with CNN Datasets CIFAR10, MNIST, SVHN | Implemented with BNN CIFAR10, MNIST, SVHN | Implemented with BNN IR dataset, MNIST |
| Network | Alexnet | Imagenet | Alexnet |
| Accuracy | 87% | 95% | 96% |
| FPS | 35 | 20 | 25 |
| Drawback | Resolution of Images are less | Edge detection of objects | Performance of IR images are less |
| Hardware | Zynq-7100 FPGA | Zynq-7100 FPGA | KintexU 115 board. |

TABLE 1: Comparison of Research Papers

**Experimental or Analytical Work Completed in the Project**

The object detection is used to detect a known class, cars and so on. An example, of a bicycle detection in an image specifies the location of certain parts. The pose is determined by a 3-Dimensional specifying the object relative to camera. The object system constructs a model from object class from object class from set of training examples. A fully connected layer can be converted by a 1-Dimensional convolutional layer. Output SoftMax layer of shape of number of classes are predicted. If we are using sliding window approach then there are passed to Binary Net four times where each time sliding window crops the input image matrix [5].

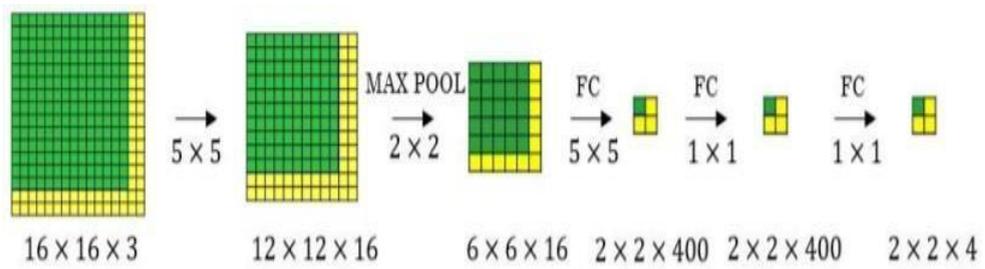

FIGURE 2: Classification of Layers

It is decided by number of filters used in Max pool layer. The main advantage is sliding window runs and computes all values simultaneously. This technique is fast, but the weakness of this techniques is that position of bounding boxes is not accurate. Googlenet is used which achieves high accuracy in running in real time. It divides the image into grids. The goal is to improve accuracy. The series of convolution feature layers are added at the end of network.

Generate the initial sub segmentation which uses generated regions to produce the final candidate region proposals. These regions are wrapped into a square and fed into BNN which consists of feature extracted from images. It takes huge amount of time to train the network. It can be implemented real time as it takes around 47 seconds for each test image [4].

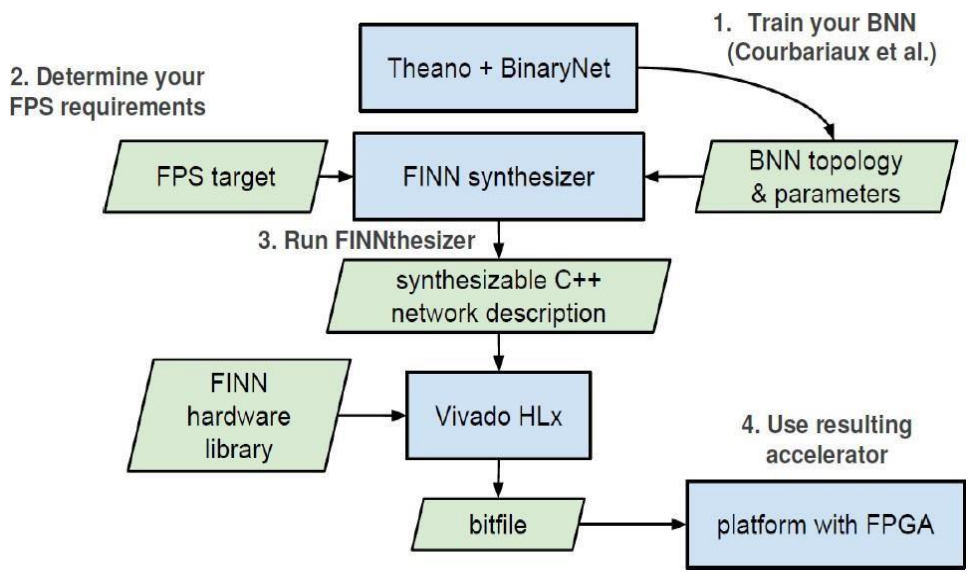

FIGURE 3: Existing Architecture for FINN

Human learners have different ways to transfer knowledge between tasks. We recognize and the relevant knowledge is applied from previous tasks to the new tasks. In development of algorithm transfer learning is the most ongoing research in deep learning. The Transfer Learning is used only to transfer knowledge that is learned and can be implemented to other models. Similarly, the last layer of CNN with the help of Googlenet is changed accordingly to improve the accuracy in identifying the objects. A Convolutional neural network (CNN) consists of one or more layers used for image processing, classification, segmentation and so on. It consists of different filters in order to improve the performance and accuracy.

The Binarized Neural Networks (BNN) is an upgradation of CNN that consists of binary weights and activations with a limit of [-1, +1] and it is easily compatible with transfer learning. CNN has less feature compatibility and it takes more time to train images when compared to BNN. First, the network structure is optimized which removes the Batch Normalization (BN) that makes BNN to run without any loss of accuracy. Second, an architecture of parallelism is proposed which is a perfect load balancer of BNN. Third, both the convolutional layers and fully connected layer are combined to give better performance in order to improve the accuracy.

The Nvidia Jetson Nano board is a GPU acts as a platform so that the proposed algorithm is implemented on it. The performance of GPU is comparatively greater and even it takes less time to show the final output. It has high throughput which helps the computer vision to identify the objects exactly.

**Convolutional Neural Network:**

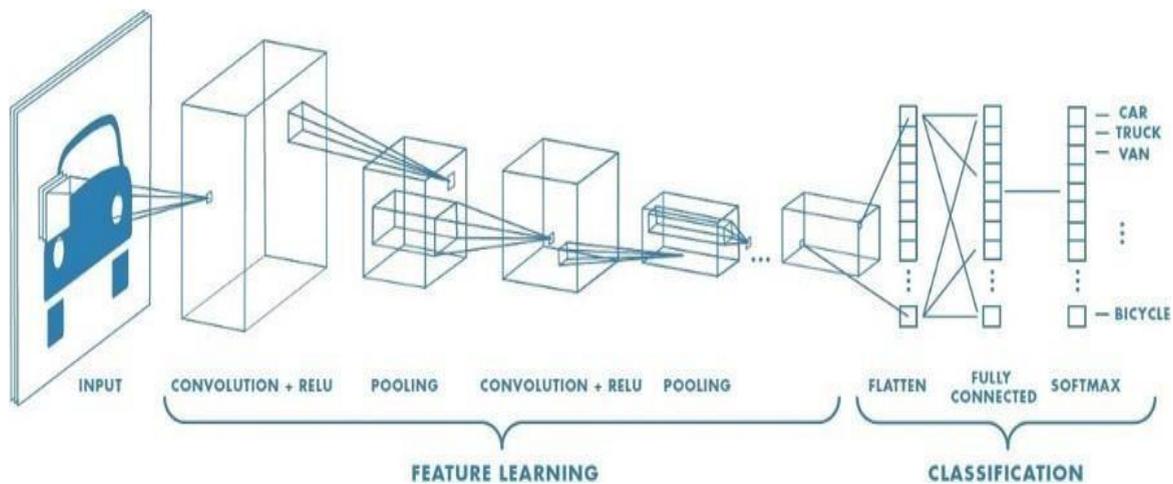

FIGURE 4: Architecture of CNN

It is a class of deep networks which is used for visual imagery. It uses multi-layer perceptron's which is a fully connected network. It can handle more complex network by simplified layers. The convolutional layers depend on convolutional kernels that are defined by width and height, and the depth of convolutional filter must be equal to number of channels of input feature map [10].

The only difference between CNN and the Artificial Neural Network (ANN) is that CNN are used in pattern recognition within the images. It allows to encode images into the architecture. ANN struggle to compute the image data this is one of the major limitations of it. MNIST data are suitable for ANN because the image dimensions are small with 784 weights. It is composed of multiple blocks such as convolution layers, pooling layers and fully connected layers which learn spatial hierarchies through back propagation algorithm.

In image pixel values are stored in a 2-Dimensional array and a kernel is applied for each position which makes CNN more efficient in image processing. Extracted features can become more complex it is called as training, which is an optimization algorithm called as backpropagation. A model under kernels and weights is of loss function which is updated through back propagation with optimization algorithm [4].

It performs down sampling of inputs that has a feature map with height×width and then further down sampled which is the average of all elements that is applied. The advantages are it reduces number of learnable parameters and enables the CNN to accept input of different sizes. Last layer activation function is applied to last fully applied layers which is different from others. Activation function is selected according to the task. It is applied to multi classification task which is a SoftMax function that is normalized and it is with a range between 0 and 1 and all values sum to 1 [7].

Finds kernels in convolution layers and weights in fully connected layers and gives the labels on a training dataset. Backpropagation is used for training neural networks which plays essential role to calculate the loss function through forward propagation by the optimization algorithm. Deep learning is used to classify images from large number of training data which uses corresponding labels for efficient classification using CNN. In object detection CNN is used to classify data based on the labeled dataset which uses unsupervised learning to the algorithm [10]. Transfer learning can be incorporated which trains a network on the large dataset such as Googlenet then it is applied to the given task. Feature extraction removes the last layer of the network and maintains the remaining network which consists of fixed feature extractor. The classifiers such as support vector machines are used which follows the fine-tuning method which is a part of kernels [11].

**Transfer Learning:**

Transfer learning is realized with the help of TensorFlow which is the artificial intelligence framework that can handle the new layers to be implemented on Googlenet. It is the improvement of learning that adds layers from the existing architecture. When network is trained on large datasets then the model is learned to perform a particular task. Fine tuning methods are obtained if the neural network is small then it is better to make changes to the last layers such that only these layers can be retrained on the network [10].

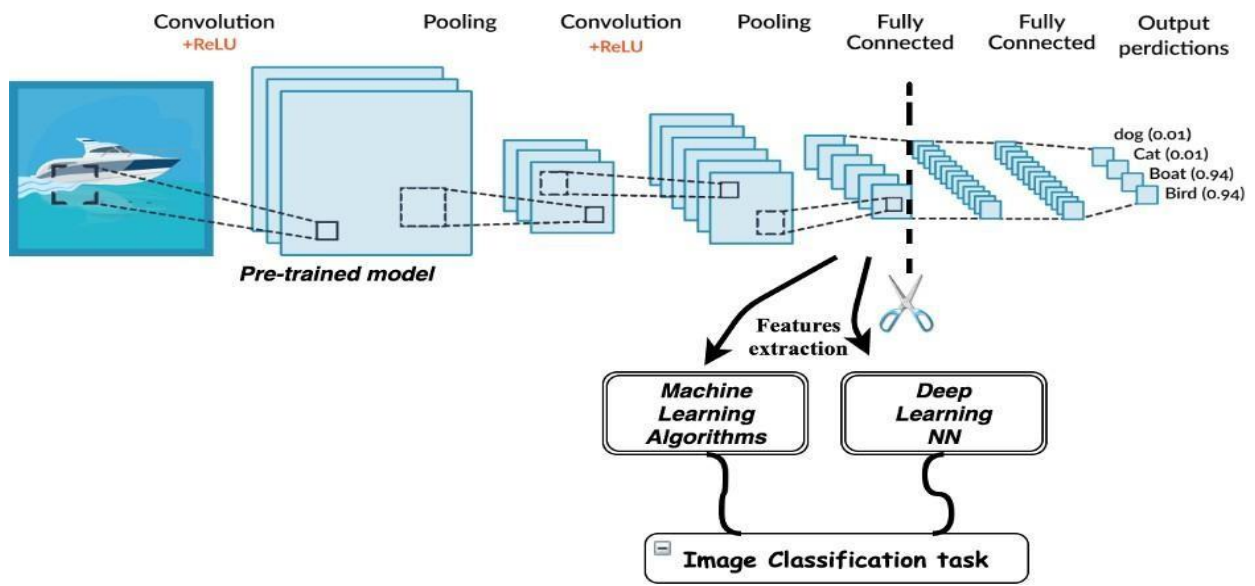

FIGURE 5: Implementing Transfer Learning on Last Layer of the network

Transfer learning is used when there is an improvement task for transfer of knowledge that has already been learned. Most algorithms have developed to assign a single task which is used to formulate and solve the challenges. Both inductive and reinforcement learning are used for task mapping. It is inspired by the human learners that they transfer knowledge between tasks. Machine learning algorithms address the isolated tasks.

There are three ways to improve transfer learning. First is initial performance in target task using transfer knowledge which is compared to initial performance when further learning is not done. Second is the amount of time that the model will take to fully learn the targeted task when transfer knowledge is compared to time which learns from scratch. Third is final performance level that is achieved in target task when compared to final level without transfer. If the performance of the model is decreased then negative transfer has occurred [10].

**Googlenet:**

It is a pre trained model that is of 22 layers which consists of nearly 1000 classes. Inception layer covers a big area but also keeps the fine resolution of images. It convolves parallel layers which is 1x1 and 5x5 matrix. The Gabor filters is a linear filter for texture analysis handles better model sizes. Googlenet uses 9 Inception layers. The final con- cat layer is obtained by pooling method. Pooling reduces the network sizes by SoftMax techniques through matrix multiplication operations. Max Pooling is the down sample of the input image whereas the average pooling is the average output of each feature maps [7].

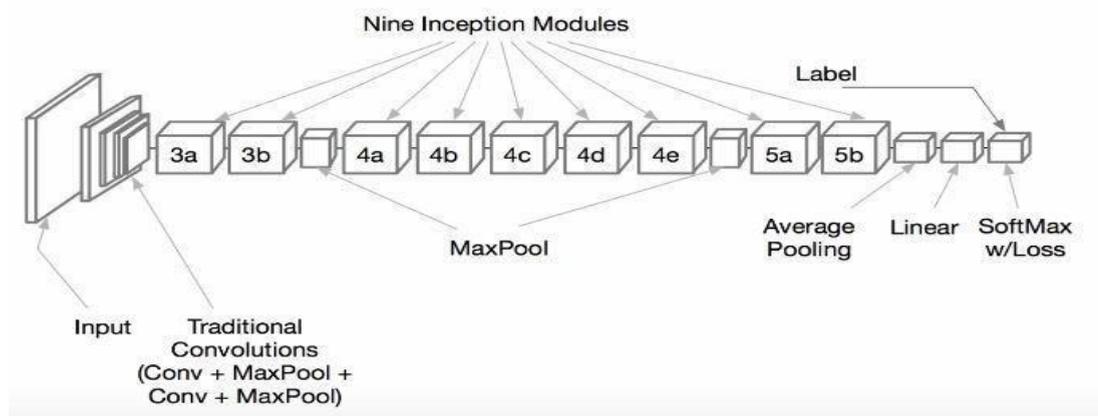

FIGURE 6: Layers of Googlenet

The noise is removed by the wiener filter and the performance of GoogLeNet is better than AlexNet on various parameter including time, accuracy. The architecture is of different size filters for same images and obtains the output. It is of 22 layers. The 1x1 convolution is used for dimension reduction. and the weights are selected with appropriate features. It contains multiple inception modules to form a deep network from which high accuracy is obtained. Transfer learning is used to recognize the objects using the numerous images and labeling of objects are based on probabilities [8].

To classify the images dataset of different images are loaded. It is of training the data and the other is validation of images. The Googlenet helps to analyze network which helps in layer information of network which gives information on weights, bias for convolutional layers. Loading pretrain network is an extraction of features from input images from convolutional layers and compared with the output layers into predicted label, loss value. Googlenet can also be used to conduct retraining of new data set. The network that is to be trained is at augmented image datastore. It restricts overfitting with a good training accuracy [9].

**TensorFlow:**

It is an open source library for dataflow which uses math library for machine learning application such as neural networks. TensorFlow is developed by Google Brain team which can run multiple GPU. The architecture is easily compatible with various plat- forms such as desktops and clusters of servers to mobile devices. The computations are in the form of dataflow graphs which performs on multidimensional arrays referred as tensors. TensorFlow is available in 64-bit Mac, Linux and Windows. Kubeflow is introduced which is the Kubernetes deployed in Tensor processing unit. In 2017 TensorFlow lite was introduced which is the GPU inference engine for micro controllers and mobile devices [12].

Keras is an open source library for neural network that can be run on top of the Tensor- Flow which acts as an interface rather than standalone framework. It is used to build blocks such as layers, objectives and activation functions which makes image data easier to analyze with the code.

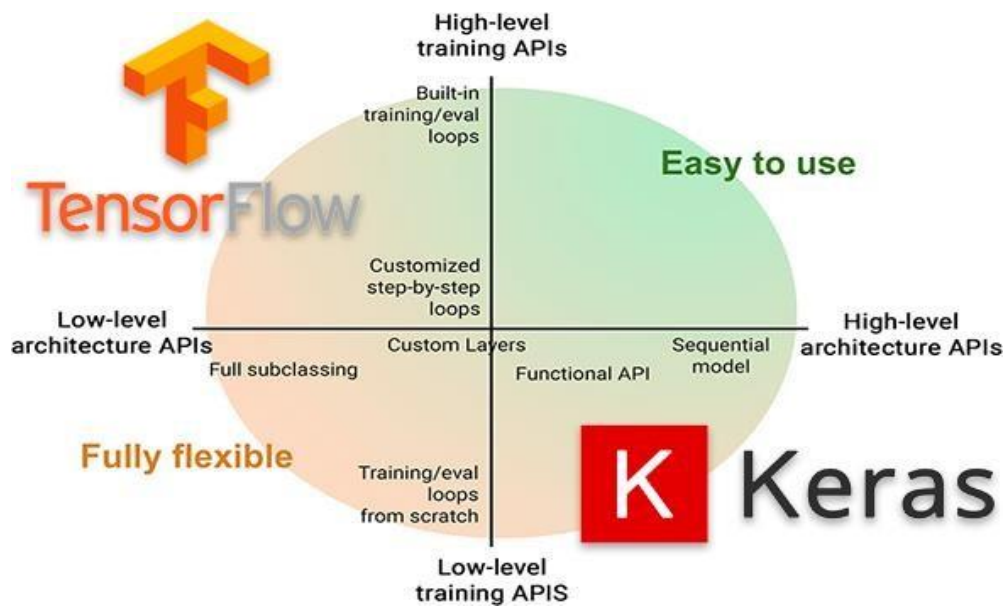

FIGURE 7: TensorFlow with Keras Package

## DESIGN / PROTOTYPE

Most of algorithm for object detection are sensitive to background and cannot detect edge of the object. It uses shape fragment features which is a multi-scale. Most methods can be classified as point-based approaches and boundary curve approaches is always limited to noises in the images. It is of pure handwritten digits. Shape based methods are segmentation errors for object detection and recognition. Appearance based methods consider the appearance feature of the object instead of geometrical properties. It uses extrapolating known strategies from 2D single view object class detection by combining classifiers for separate viewpoints. To match the shape robustly, iterative methods are used. It contains one key point and two connected line segments. For a rigid object, if more than 3 points are matched then it is easy to estimate the object pose. When view point changes appearance of object is different. For 2D object modeling we need to setup the constraint between all the views for same object. For 3D object modeling it is a model that is a Multiview representation of a 3D object class [13].

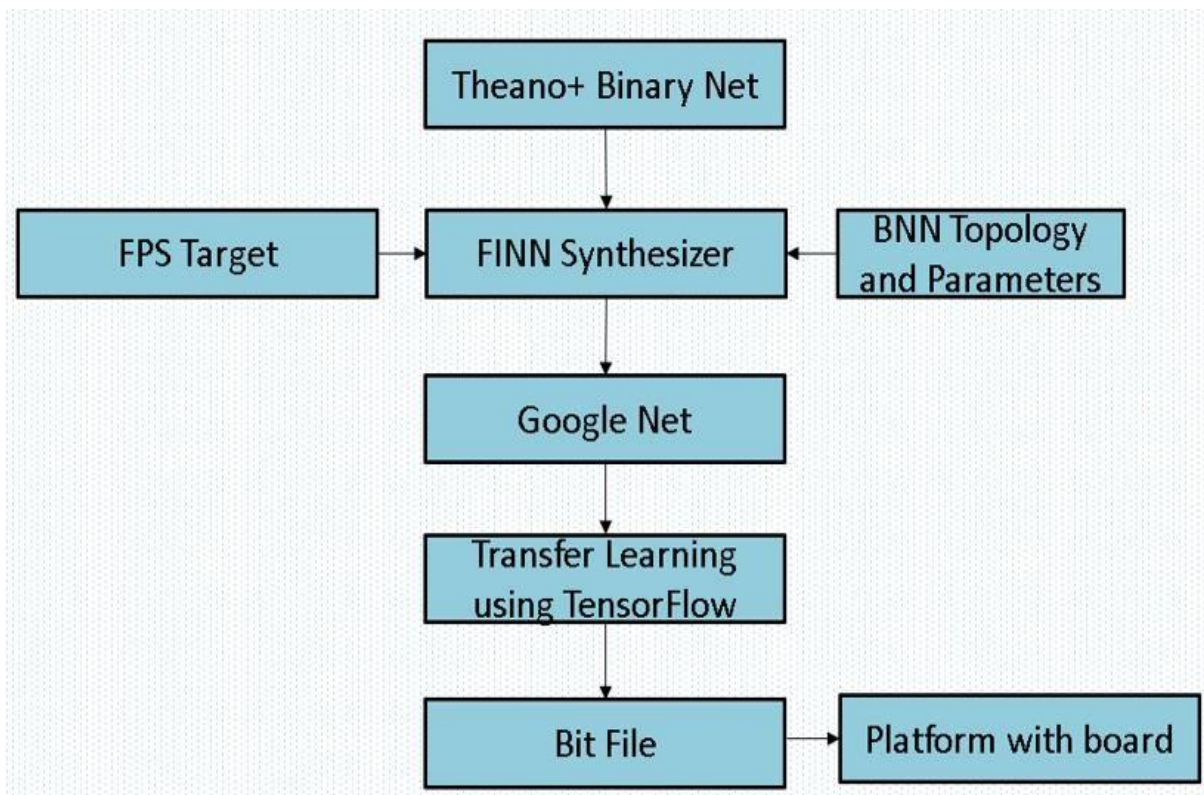

FIGURE 8: Proposed Architecture

To build a custom architecture rather than the fixed architecture were all neural net- works are kept in on chip memory. This avoids the off-chip memory which minimizes the latency by overlapping the computation. The separate mapping of layers ensures that it has heterogeneity.

A BNN accelerator has various constraints when imposed on the use case. It uses clock frequency to control the classification throughput. BNN uses 1-bit values for all input activations, weights and output activations which computes binary dot products with fewer hardware resources. The hardware can be implemented by the pop count operation that counts number of set bits in batch normalization which determines the output activation. The block diagram consists of Theano, Binarynet, BNN. FPS target, Finn synthesizer, Googlenet, Transfer learning, bit file and platform with board.

**Hardware Implementation:**

Nvidia Jetson Nano is used to run multiple neural networks for image classification, object detection. It is basically implemented by Baleno Etcher which is installed in the SD Card. GPU is used for parallel operations mainly for fast computing and a large dataset can be processed easily.

The GPU is most preferred when compared to CPU. The GPU have high computedensity and high computations per Memory access. It is mainly built for parallel operations which includes many parallel execution units (ALUs) and Graphics is the best case of parallelism. Deep pipelines are used which is of hundreds of stages with high through- put and latency [8].
It has better flow control logic, gather memory access. The CPU have low compute density and complex control logic with larger caches. It is optimized for serial operations with fewer execution units (ALU) and higher clock speeds. It consists of shallow pipelines which is of less than 30 stages and it includes low latency tolerance whereas new CPU have more parallelism. The components are Micro SD Card slot, 40 pin expansion header, Micro USB Port gigabit Ethernet Port, USB 3.0 ports, HDMI output port, display port connector DC Barrel jack , MIPI CSI Camera connector [10].

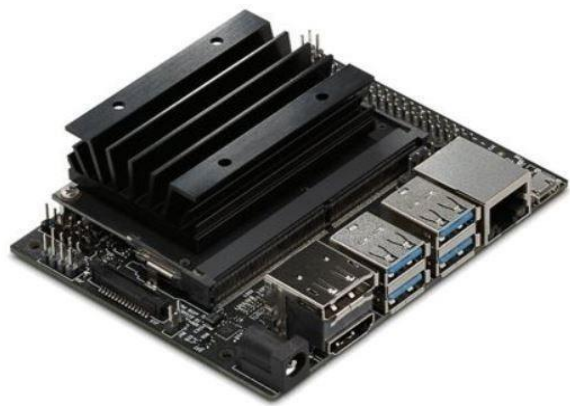

FIGURE 9: Physical Structure of board

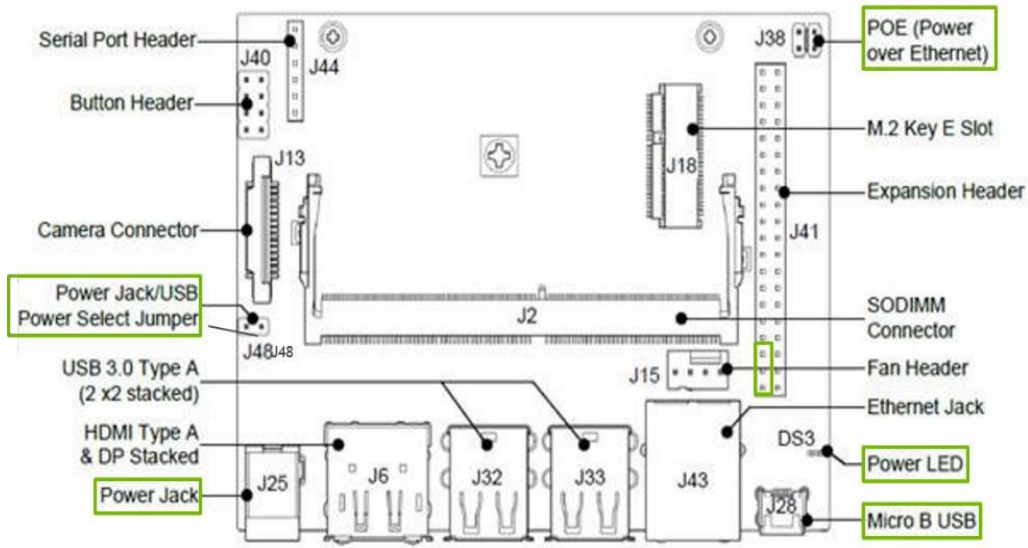
FIGURE 10: Components of Nvidia Jetson Nano

Googlenet retrain these models which often choose between fast processing and prediction accuracy. Object detection is chosen over image classification because of determining size and position of objects in images. It focuses on retraining pre trained models with custom datasets with known framework. Classifies the images through retraining the model by transfer learning mainly uses TensorFlow object detection API framework. It is trained on the inception v2 dataset. Cameras capture different viewing angles of prototype. To increase the stability of models when training, evaluating and testing all models. The training model uses Googlenet framework with set of 0.001 and a batch size of 64 [13].

# RESULTS AND DISCUSSION

It gives a detailed review on deep learning, based object detection frameworks which handles different clutter and low resolution with modifications using BNN. Then the transfer learning and the Googlenet are compared [14].

For the purpose of this project the available MNIST, CIFAR-10, SVHN dataset is used. It consists of 25k object annotations. The project is implemented in Python 3. Googlenet is used for image pre-processing. Fine tuning is done which is trained end to end on the dataset [12].

| Methods | Accuracy | | | | | | | | | | FPS | Test Time |
|---|---|---|---|---|---|---|---|---|---|---|---|---|
| | Bottle | Chair | Person | Book | Laptop | Keyboard | Mouse | Bowl | Remote | Mobile | | |
| Previous Paper | 67 | 58 | 72 | 63 | 75 | 86 | 72 | 59 | 68 | 72 | 45 | 3s |
| GoogLeNet without TL | 70 | 60 | 75 | 65 | 80 | 90 | 80 | 60 | 70 | 70 | 50 | 3s |
| GoogLeNet with TL | 85 | 93 | 94 | 85 | 92 | 93 | 96 | 75 | 95 | 90 | 18 | 4s |

TABLE 2: Comparison of accuracy of objects

Object detection plays a vital role and it has to be accurate with a very less margin of errors with the least amount of resources possible. In this project few of the image detection algorithms which is used to develop a system to detect objects mainly used in the automation industry [5]. By this the accuracy of Googlenet with Transfer Learning is better than the Googlenet.

These solutions take multiple regions of the frame for feature extraction and perform the object detection over these regions [6]. The objects are stored in the database and are compared with the objects. The system identifies object based on the threshold value example: 35% feature mapping criteria and the objects are identified from database [9].

A particular class of such methods relies on graph models where an object is decom- posed into a number of parts, each one is represented by a graph vertex. The efficient modelling as graphs can be used for effective representation of images for detection and retrieval of objects. The instruments capable to classify such huge amounts of data has become a real challenge [2].

Googlenet provides features for customization, and production capable deployments for object detection tasks. Adjusting minimum probability by default, objects detectedless than 50% will not be shown.

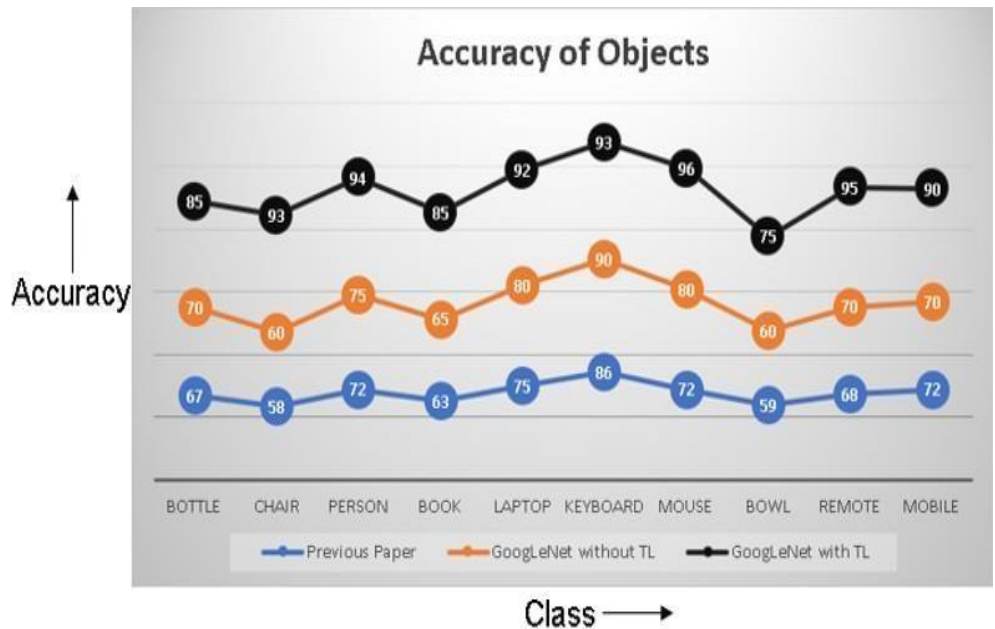
FIGURE 11: Graph of Accuracy v/s Class

BNN topology follows the backpropagation model which shows greater accuracy when it is combined with the neural networks like Googlenet. Model sizes are much smaller than the full precision model. It consists of unseen topologies, traffic matrices and con- figurations that are used as the baseline of BNN. At digital design the inference time is less at BNN and even execution is faster and accurate. The bit width frequency is reduced which is the off-chip data transfer and the overhead of bits. FINN synthesizer has a fast, scalable inference that has a significant redundancy and high classification accuracy which reduces the floating point to binary values. It uses flexible heterogeneous architecture which comprises of convolutional and pooling layers that classifies billions of images with least possible time.

Googlenet is a pretrained network that is 22 layers deep which classifies nearly 1000 objects like keyboard, mouse, laptop and so on. It contains pooling filter and activation functions. The pooling is of two types, one is max pooling which is the down sample of the input and another is the average pooling which is the average output of each feature map. Pooling is the spatial reduction of network filters that uses SoftMax technique which is the normalized exponential function. The main reason Googlenet is chosen in the model is that the network size is less so that more images can be trained in the dataset. Transfer learning is the transfer of knowledge of the model that can be incorporated to similar models to obtain a better output.

This can be increased for high certainty cases. Using a custom object class, we can find the object class of unique objects. Detection speed can be reduced by setting the speed to be fastest [8]. Input type is the file path to an image file stream of an image as input image. Output type specify the detect object from image function should return image in the form of a file.

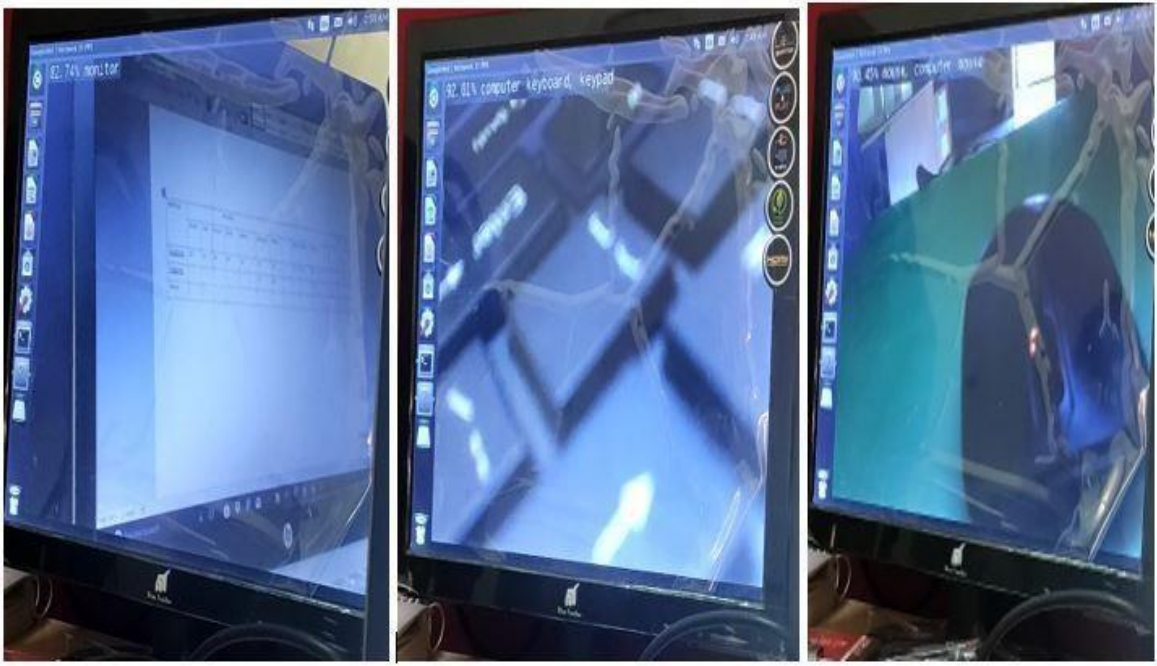

FIGURE 12: Googlenet

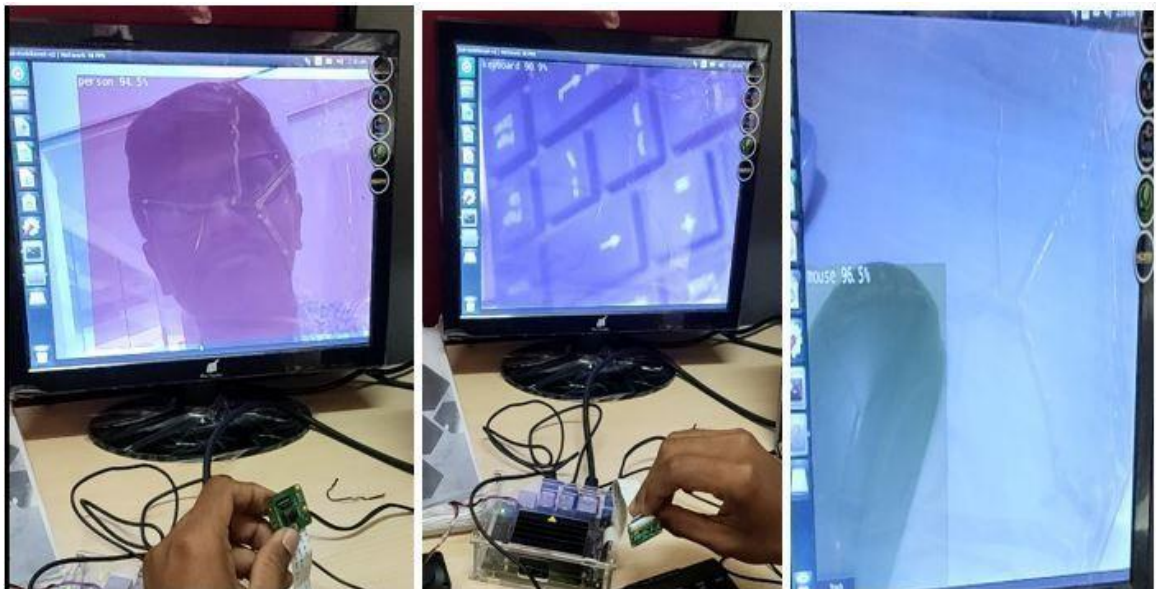

FIGURE 13: Googlenet with Transfer Learning

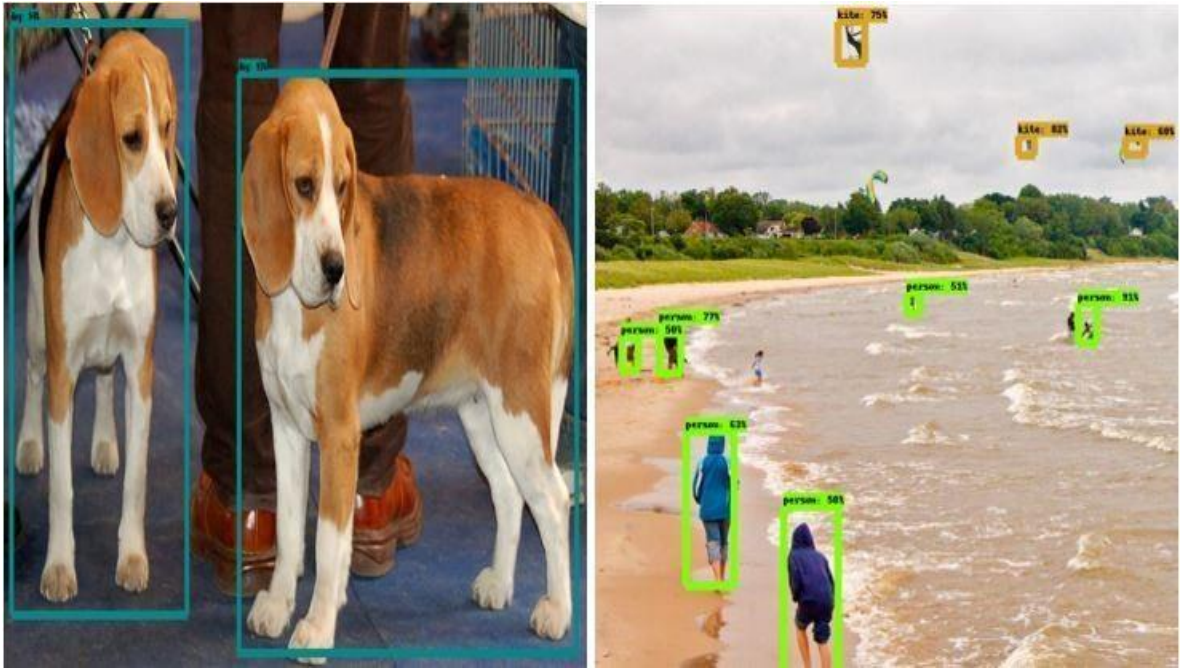

FIGURE 14: Before Transfer Learning is applied on Pre trained Images

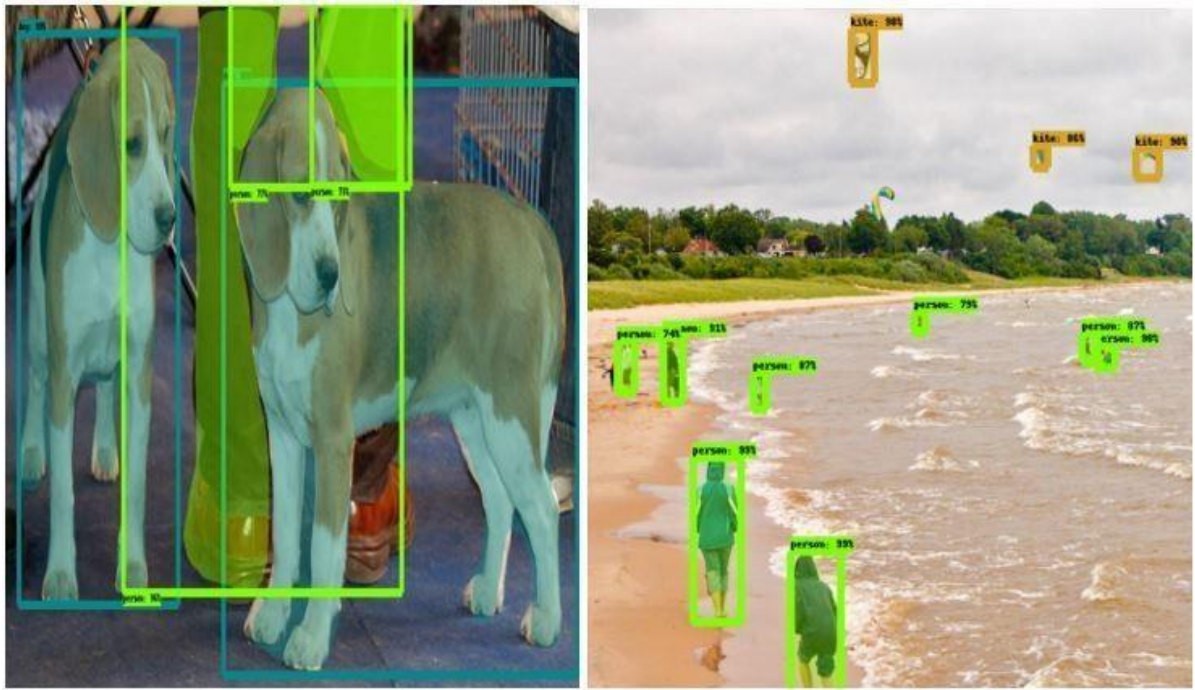

FIGURE 15: After Transfer Learning is applied on Pre trained Images

# CONCLUSION

An accurate and efficient object detection algorithm is developed and studied using BNN with transfer learning technique. This project uses recent techniques in the field of computer vision and deep learning. It can be used in real time to determine accuracy of objects. The Googlenet has an average precision of 93% whereas Googlenet with Transfer Learning it has reported an average precision of 98% for different objects. It is also flexible to extend to other frameworks. The Tensor Flow API is used to determine the accuracy of a object from the images present in the dataset. It shows the comparison of the objects before and after where accuracy is improved drastically. It overall creates the instances of objects from particular class in an image which classifies the object. A high performance is obtained by using the Tensor Flow API which has minimal power consumption and latency. This technology is mainly used for embedded applications and automation industry. Nvidia Jetson Nano has the good performance were this hard- ware platform is constantly updated by deep neural network libraries which increases the performance in computer vision tasks.

As a step for future work, several short coming that came across during the development of the proposed algorithm are analyzed in some extent. During detection process we don't know whether to detect object first or the part first. Using depth images, it is easy to segment the objects using thermal camera. For properly detecting the background objects, a 3D model can be proposed to point object detection and segmentation of images. Further it can be focused on implementing larger networks like Resnet and higher performance convolutions can be obtained.